\documentclass[letterpaper, 10 pt, conference]{ieeeconf}
\IEEEoverridecommandlockouts
\usepackage{amsmath,amssymb,amsfonts,bm}
\usepackage{algorithmic}
\usepackage{graphicx}
\usepackage{textcomp}
\usepackage{xcolor}
\usepackage{lipsum}
\usepackage{balance}
\usepackage[style=ieee,dashed=false,isbn=false,url=false,doi=false]{biblatex}
\def\BibTeX{{\rm B\kern-.05em{\sc i\kern-.025em b}\kern-.08em
    T\kern-.1667em\lower.7ex\hbox{E}\kern-.125emX}}

\begin{document}

\title{Optimal Trajectory Planning in a Vertically \\Undulating Snake Locomotion using Contact-implicit Optimization}

\author{
Adarsh Salagame$^{1}$, Eric Sihite$^{2}$, Alireza Ramezani$^{1*}$% <-this % stops a space
\thanks{$^{1}$This author is with the Department of Electrical and Computer Engineering, Northeastern University, Boston MA
        {\tt\small salagame.a, a.ramezani@northeastern.edu*}}%
\thanks{$^{2}$ This author is with California Institute of Technology, Pasadena CA
		{\tt\small esihite@caltech.edu}}%
\thanks{$*$Indicates the corresponding author.}
}
\maketitle

\begin{abstract}

Contact-rich problems, such as snake robot locomotion, offer unexplored yet rich opportunities for optimization-based trajectory and acyclic contact planning. So far, a substantial body of control research has focused on emulating snake locomotion and replicating its distinctive movement patterns using shape functions that either ignore the complexity of interactions or focus on complex interactions with matter (e.g., burrowing movements). However, models and control frameworks that lie in between these two paradigms and are based on simple, fundamental rigid body dynamics, which alleviate the challenging contact and control allocation problems in snake locomotion, remain absent. This work makes meaningful contributions, substantiated by simulations and experiments, in the following directions: 1) introducing a reduced-order model based on Moreau's stepping-forward approach from differential inclusion mathematics, 2) verifying model accuracy, 3) experimental validation.

\end{abstract}

\section{Introduction}

Contact-rich problems such as snake robot locomotion offer unexplored, rich optimization-based trajectory and acyclic contact planning opportunities, which can largely have a positive impact on other applications—such as prehensile/nonprehensile multi-finger object manipulation \cite{jiang_contact-aware_2023,woodruff_planning_2017,chavan-dafle_stable_2018,xiao_one-finger_2024}--by providing efficient controllers for fast and dexterous manipulation (thanks to the inherent connectivity between locomotion and manipulation).

Trajectory and acyclic contact planning in robots that intermittently interact with their surrounding world have been studied extensively in recent years. However, the problem posed by snake locomotion is more sophisticated because (1) manifold contact points are involved and (2) the action space is high-dimensional, which increases the number of possible control actions. Consequently, control design in snake locomotion predominantly concerns contact or control allocation.

\begin{figure}[t]
    \centering
    \hfill % Pushes figure to the right
    \includegraphics[width=1\linewidth]{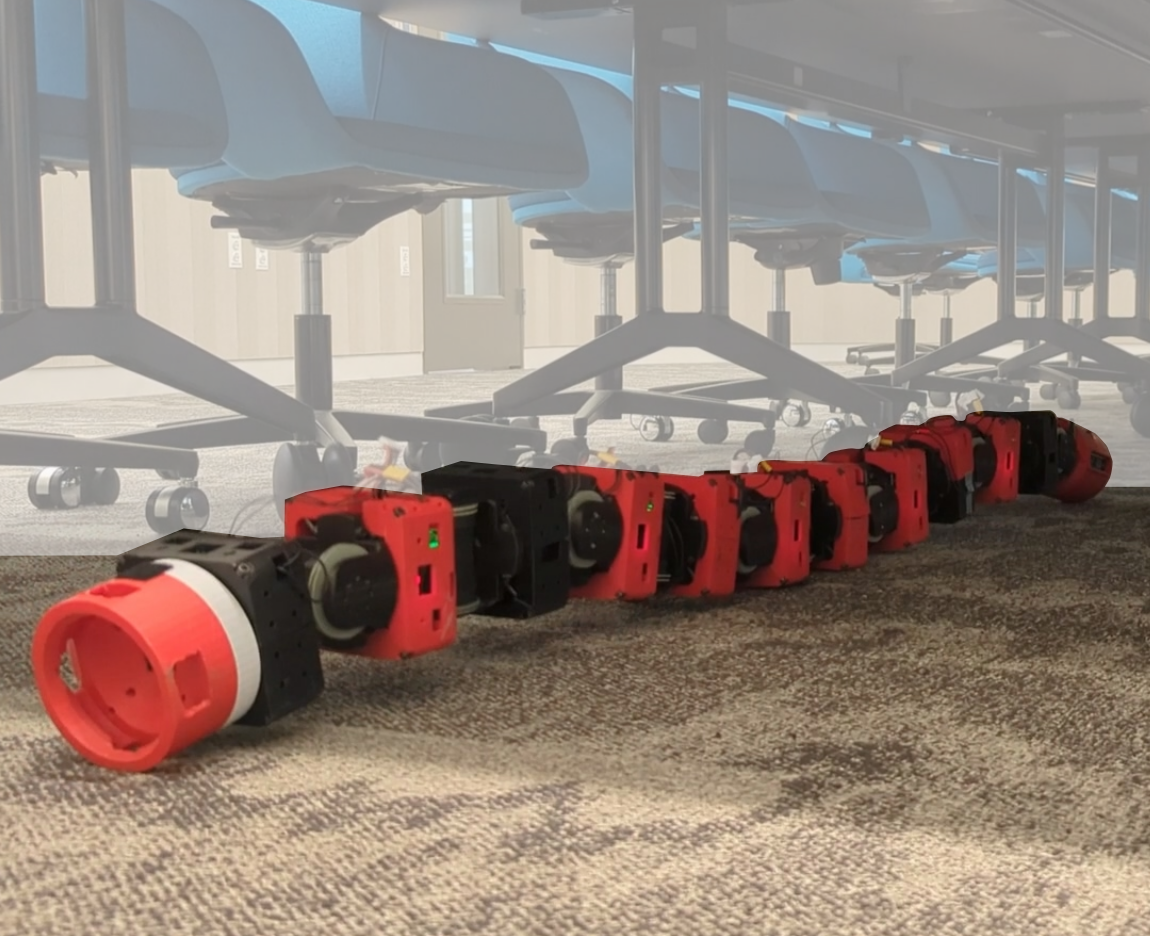}
    \caption{Shows COBRA performing vertical undulation for traveling through a confined space.}
    \label{fig:cover}
\end{figure}

So far, a rich body of control work has been developed, including concepts based on the backbone curve method \cite{huang_unified_2024}, floating frame of reference \cite{melo_modular_2015}, virtual chassis \cite{rollinson_virtual_2012}, and the use of torsion and curvature to describe snake shape \cite{melo_modular_2015}, along with economic Model Predictive Control (MPC) \cite{nonhoff_economic_2019}, MPC for viscous environments \cite{hannigan_automatic_2020}, and the Frenet-Serret framework \cite{yamada_study_2006}. While the main focus of snake robotics research has been on emulating snake locomotion and replicating its distinctive movement patterns using central pattern generators, models and control frameworks that alleviate the challenging contact and control allocation problems faced in snake locomotion remain absent.

Mathematical models of rigid body interactions based on differential inclusion \cite{moreau_unilateral_1988} can capture the dominant dynamics of these systems. These models--founded on implicitly resolving stepping forward steps in the system states, instead of the typical Euler approach that follows a fixed-point gradient fashion $x_{k+1} = E(x_k)$ used in no-contact rigid-body models--allow modeling contact kinematics and forces as nonlinear functions of joint actions. These functions can then be used to reformulate the control problem into a proximal optimization problem for which many solution tools exist.

Contact-implicit optimization planning forms a backbone research component in many domains of robot locomotion \cite{pitroda_quadratic_2024,krishnamurthy_thruster-assisted_2024,sihite_posture_2024}. The key question is: how can we revisit these tools in the context of snake locomotion? In this regard, leveraging models of snake locomotion that use simplified rigid contact models, although not fully capturing the complexity of a snake’s contact-rich locomotion, can enable the control of these simplified models in most practical scenarios.

If successful, this approach directly connects snake locomotion to the rich body of work already underway in multi-contact manipulation and locomotion, and it allows for the control of the full-dynamics a snake exhibits through optimal joint movements that respect cone and complementarity conditions. We can then leverage the inherent redundancy of the snake body and reformulate snake locomotion as an acyclic self-manipulation problem.

This work makes the following contributions: 1) introducing a reduced-order model based on Moreau's stepping-forward approach from differential inclusion mathematics, 2) verifying model accuracy, 3) experimental validation. The paper is organized as follows: In Section \ref{sec:overview} we present an overview of the COBRA platform upon which this work is based. In Section \ref{sec:model} we discuss the dynamic modeling and proximal optimization framework for the reduced order model. In Section \ref{sec:results} we show experimental and simulation results that show model agreement between the analytical model and experiment. Finally in Section \ref{sec:conclusion} we end with concluding remarks and future directions.

% \begin{itemize}
%     \item introduces a reduced-order model
%     \item verifies the accuracy of the model 
%     \item experimental validation
% \end{itemize}

% In this paper, ...

% \begin{itemize}
%     \item utilize cobra
%     \item focus on vertical undulation gait on flat ground
%     \item nonprehensile manipulation of a box
%     \item assume virtual spherical convex sets encapsulate robot modules
%     \item these convex sets form gap functions for Moreau's implicit marching forward approach
%     \item control actions obtained during stepping forward as a SOCP
% \end{itemize}

\section{Overview of Hardware Platform}
\label{sec:overview}
\begin{figure}
    \centering
    \includegraphics[width=1\linewidth]{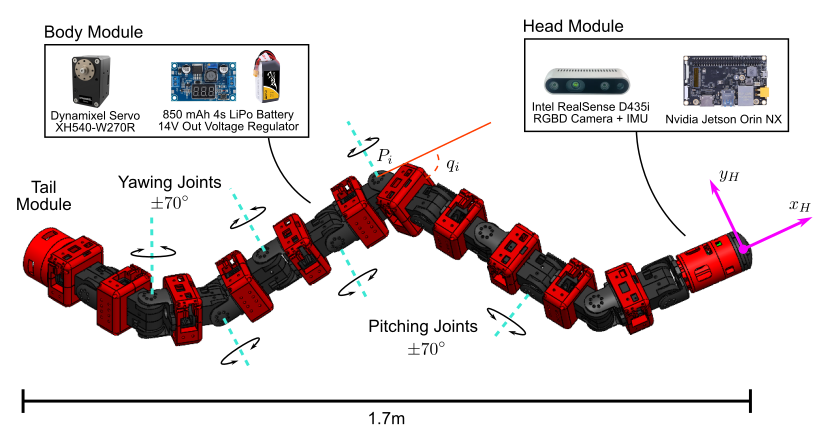}
    \caption{COBRA system overview.}
    \label{fig:sys-overview}
\end{figure}

COBRA \cite{salagame_loco-manipulation_2024, salagame_dynamic_2024, jiang_hierarchical_2024-1, salagame_how_2024, salagame_non-impulsive_2024, salagame_heading_2024, jiang_snake_2024} is a snake robot composed of twelve interconnected links and eleven actuated joints arranged with alternating axes of rotation (Figure~\ref{fig:sys-overview}). Each identical body module contains a Dynamixel XH540-W270-R servo, a 4S 850,mAh LiPo battery, and a voltage regulator, all daisy-chained via an RS-485 bus. An NVIDIA Jetson Orin NX (8GB RAM), housed in the head module, centrally controls these actuators. The head module also integrates an Intel RealSense D435i stereo camera and IMU for perception, while the tail module incorporates a customizable payload compartment for task-specific electronics.

Within the head module, the Jetson Orin NX executes a 500 Hz control loop that sends position commands to the servos and reads back proprioceptive data, including position, velocity, and current (torque sensing and controls). The RealSense D435i provides synchronized RGB, stereo depth, and inertial measurements, enabling both low-level control and higher-level autonomy tasks such as mapping and navigation. 

\section{Modeling}
\label{sec:model}
\begin{figure*}
    \centering
    \includegraphics[width=0.83\linewidth]{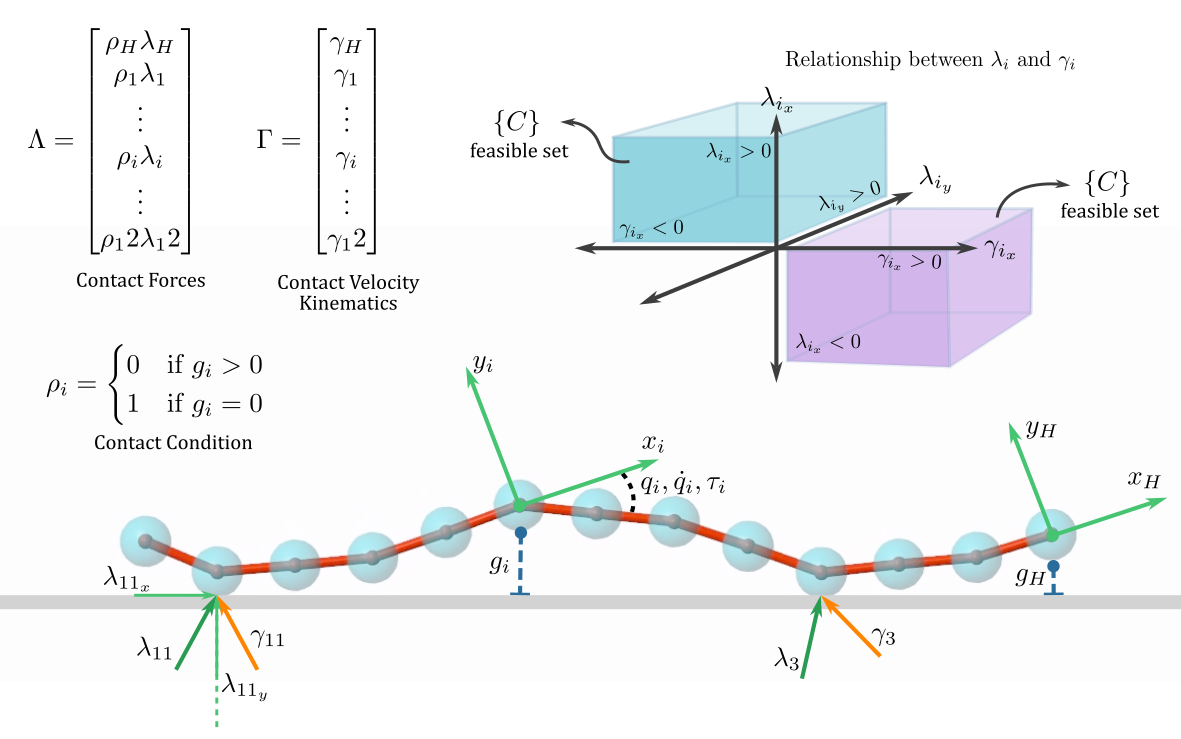}
    \caption{Illustration of our modeling and control approach based on proximal optimization for the constrained COBRA model with contact dynamics, where sliding between rigid bodies is allowed.}
    \label{fig:fbd}
\end{figure*}

We assume a snake model on flat ground, as shown in Figure~\ref{fig:fbd}. The robot's interactions with the environment are modeled based on rigid body dynamics by assuming virtual convex objects (spheres) that encapsulate each module. These convex sets then form gap functions that are used to integrate the state trajectories in the resulting differential inclusion problem.
    
The equations of motion for the system are given by: 
\begin{equation}
M(q) \ddot{q} + h(q, \dot{q}) = S u + \sum_{i} J_{c,i}^T(q) f_{c,i}+W(q)w
\end{equation}
where \( q \) represents the configuration of the robot, and \( \ddot{q} \) is its acceleration. The mass matrix \( M(q) \) depends on the configuration. The term \( h(q, \dot{q}) \) includes Coriolis, centrifugal, and gravitational effects. The actuation matrix \( S \) maps the control inputs \( u \) into the system, while \( W(q) \) maps the external wrenches \( w \) from robot-object interactions. The term \( \sum_{i} J_{c,i}^T(q) f_{c,i} \) represents the forces applied at contact points, where \( J_{c,i}(q) \) is the contact Jacobian and \( f_{c,i} \) is the contact force.

While one can argue that the typical quasi-static motions showcased by snake robots justifies simpler models, we have decided to present a complete full-dynamics of our system. This way, we can capture nonlinear behaviors such as inertial dynamics contributions during dynamic locomotion or loco-manipulation scenarios. The contact force \( f_c \) is constrained by the Coulomb friction model,  
\begin{equation}
f_c \in \{ f \mid f_n \leq 0, \quad \| f_t \| \leq \mu | f_n | \}
\end{equation}
where \( f_n \) is the normal force component and must be compressive (\( f_n \leq 0 \)), \( f_t \) is the tangential force, and \( \mu \) is the coefficient of friction. Instead of imposing a strict no-slip condition, the model treats contact constraints as a differential inclusion:  
\begin{equation}
\begin{aligned}
    M(q) \ddot{q} + h(q, \dot{q}) - S u - W(q) w &\in \operatorname{Range}\left\{J_c^T(q)\right\}, \\\quad f_c \in F(J_c(q) \dot{q})
\end{aligned}
\end{equation}
where \(J_c=[J_{c,1}, \dots,J_{c,13}]\) which allows for both sticking and slipping conditions.  

To handle contact events, the Moreau–Jean time-stepping scheme is used. Over the interval from \( t_n \) to \( t_{n+1} \), the first step is to predict an unconstrained velocity ignoring contact forces:  
\begin{equation}
\tilde{v} = v_n + \Delta t M^{-1} [S u_n + W(q_n) w_n - h(q_n, v_n)]
\end{equation}
Then, a correction is applied via a reaction \( R \) to enforce the contact constraints:  
\begin{equation}
v_{n+1} = \tilde{v} + M^{-1} R
\end{equation}
The updated position follows a semi-implicit integration:  
\begin{equation}
q_{n+1} = q_n + \Delta t v_{n+1}
\end{equation}
Below, we provide a more detailed discussion on how $R$ is obtained.

\subsection*{Moreau's Stepping Forward Approach} 
For resolving contact forces, a SOCP is formulated. The state before contact is given by \( q_n \in \mathbb{R}^{12} \) and \( v_n = \dot{q}_n \in \mathbb{R}^{12} \), and the mass matrix at that state is \( M_n = M(q_n) \). Known forces include \( h_n = h(q_n, v_n) \), \( S u_n \), and \( W(q_n) w_n \). The free velocity $\tilde{v}$, ignoring the contact dynamics, is  
\begin{equation}
\tilde{v} = v_n + \Delta t M_n^{-1} (S u_n + W(q_n) w_n - h_n)
\end{equation}
For each contact \( i = 1, \dots, c \), the following quantities are defined: 
\begin{itemize}
    \item the contact normal \( n_i=[0,1] \in \mathbb{R}^2 \) (2D case), 
    \item the friction coefficient \( \mu_i \geq 0 \), 
    \item and the contact position and velocity \( (p^n_i, \gamma_i=\dot{p}^n_i) \). 
\end{itemize}
The decision variables include the normal contact force \( f^n_{c,i} \geq 0 \), the tangential impulse \( f^t_{c,i} \in \mathbb{R} \), and the normal gap \( g_i \geq 0 \). The velocity $v_{n+1}$ after applying the contact impulses is  
\begin{equation}
v_{n+1} = \tilde{v} + \Delta t M_n^{-1} \sum_{i=1}^{c} J_{c,i}^T [f^n_{c,i},f^t_{c,i}]^T
\end{equation}
The normal contact constraints ensure nonpenetration and enforce complementarity conditions:  
\begin{equation}
g_i \geq 0, \quad f^n_{c,i} \geq 0, \quad g_i f^n_{c,i} = 0
\end{equation}
The normal gap evolution follows a discrete update rule:  
\begin{equation}
g_i = g_i^n + \Delta t n_i^T J_{c,i} (v_{n+1})
\end{equation}
which ensures that if the contact force is active, the gap remains zero. The friction cone constraint for tangential forces is enforced as  

\begin{equation}
|f^t_{c,i} | \leq \mu_i |f^n_{c,i}|
\end{equation}

The complete SOCP formulation consists of minimizing a small regularization cost \( \Phi(v_{n+1}) \), subject to the above constraints, ensuring that all contacts and post-contact states \( (q_{n+1}, v_{n+1}) \) are solved consistently. Finally, the position update follows  
\begin{equation}
q_{n+1} = q_n + \Delta t v_{n+1}
\end{equation}
which integrates the effect of impulses and constraints into the simulation framework. We resolve state updates and obtain optimal control actions using the following optimiztion:

% \begin{align}
%     \min_{f^n_{c,i}, f^t_{c,i}, g_i, v_{n+1},u_n} \quad & \Phi(v_{n+1})=u^T_nu_n  \quad \text{(regularization term)} \notag \\
%     \text{subject to} \quad & g_i \geq 0, \\
%     &f^n_{c,i} \geq 0,\\
%     & |f^t_{c,i} |_2 \leq \mu_i |f^n_{c,i}|, \\
%     &g_i f^n_{c,i} = 0, \\
%     & q_{n+1} - q_n - \Delta t v_{n+1}=0, \\
%     & g_i - g_i^n - \Delta t n_i^T J_{c,i} (v_{n+1})=0,\\
%     & v_{n+1} - \tilde{v} - \Delta t M_n^{-1} \sum_{i=1}^{c} J_{c,i}^T [f^n_{c,i},f^t_{c,i}]^T=0.
% \end{align}
\begin{align}
    \min_{\substack{
        f^n_{c,i}, f^t_{c,i}, g_i, 
        v_{n+1}, u_n
    }} \quad & \Phi(v_{n+1}) = u^T_n u_n  \quad \text{(regularization term)} \notag \\
    \text{subject to} \quad & g_i \geq 0, \\
    & f^n_{c,i} \geq 0, \\
    & |f^t_{c,i}|_2 \leq \mu_i |f^n_{c,i}|, \\
    & g_i f^n_{c,i} = 0, \\
    & q_{n+1} - q_n - \Delta t v_{n+1} = 0, \\
    & g_i - g_i^n - \Delta t n_i^T J_{c,i} (v_{n+1}) = 0, \\
    & v_{n+1} - \tilde{v} - \Delta t M_n^{-1} \sum_{i=1}^{c} J_{c,i}^T 
    \begin{bmatrix}
        f^n_{c,i} \\
        f^t_{c,i}
    \end{bmatrix} = 0.
\end{align}

% \section{Contact-implicit optimization}

% \begin{itemize}
%     \item regularization term
%     \item decision parameters: contact forces (normal and tangent), state update, joint torques
%     \item complementarity condition at contacts
%     \item second-order cone condition at contacts
%     \item free velocity update 
%     \item state update
%     \item gap function update
% \end{itemize}

\section{Results and Discussion}
\label{sec:results}

\begin{figure}
    \centering
    \includegraphics[width=1\linewidth]{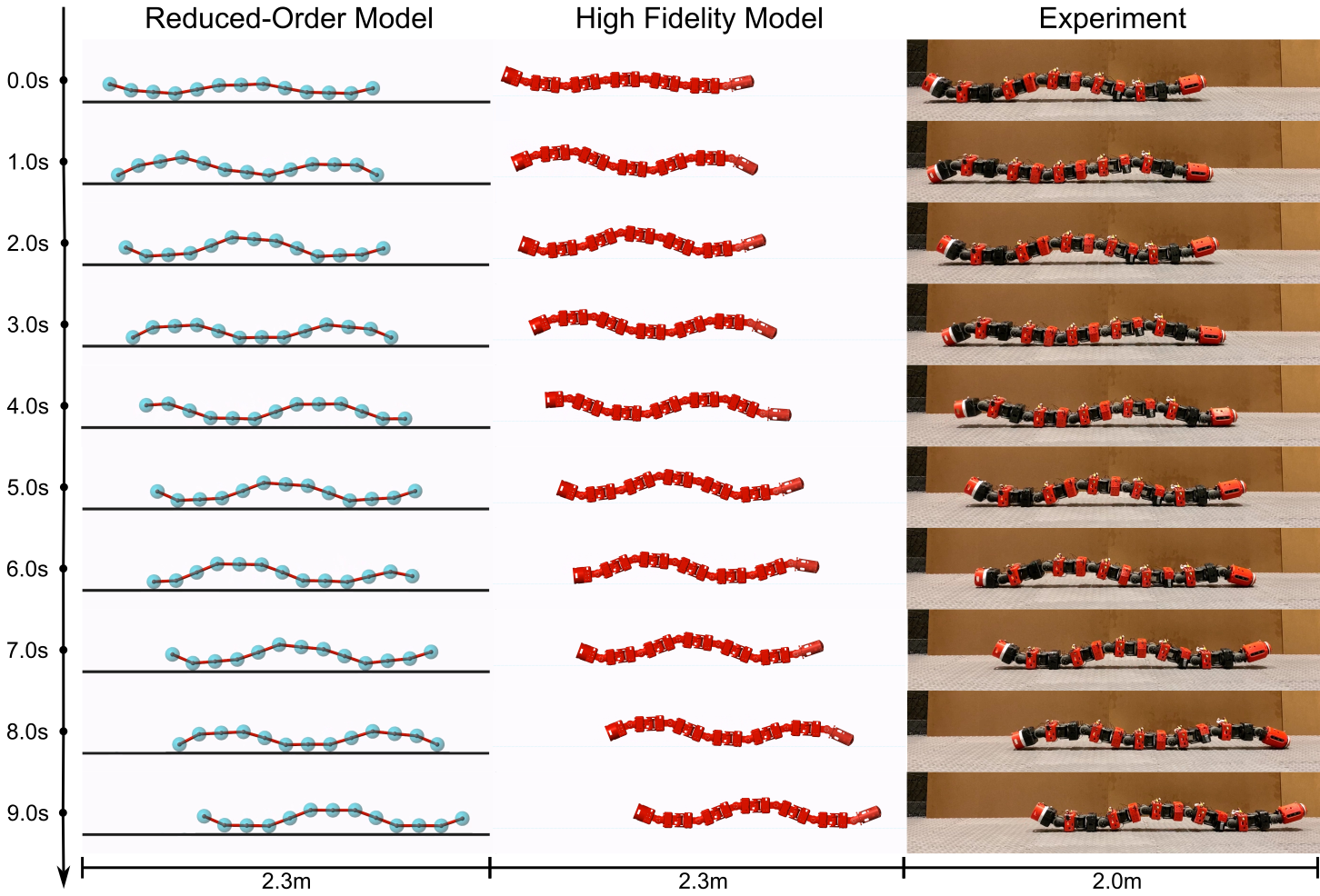}
    \caption{Shows a comparison between prediction of vertical undulation motion snapshots in experiment, high-fidelity model (Simscape) and ROM.}
    \label{fig:snapshots}
\end{figure}

The presented model was implemented in MATLAB and executed over a 10-second window to predict the robot’s behavior under a vertical undulation gait. This gait was defined by a prescribed joint trajectory, from which joint torques, contact forces, and base accelerations were computed using constrained dynamics and Moreau’s Stepping Forward Approach. Model predictions were compared against two simulation frameworks and experimental data.

A reduced-order model (ROM), depicted in Figure \ref{fig:fbd}, was implemented in MATLAB Simulink using the Simscape Multibody toolbox to study simplified contact interactions. Additionally, a high-fidelity Simulink simulation was developed, incorporating the robot’s CAD-based geometry with inertial properties computed from its shape, offering a more realistic representation of contact dynamics.

The same vertical undulation gait was executed across all three models—the ROM, the high-fidelity simulation, and analytical model, as well as on the real robot, by providing joint position commands. Their resulting trajectories, joint torques, and contact forces were analyzed.

\begin{figure}
    \centering
    \includegraphics[width=1\linewidth]{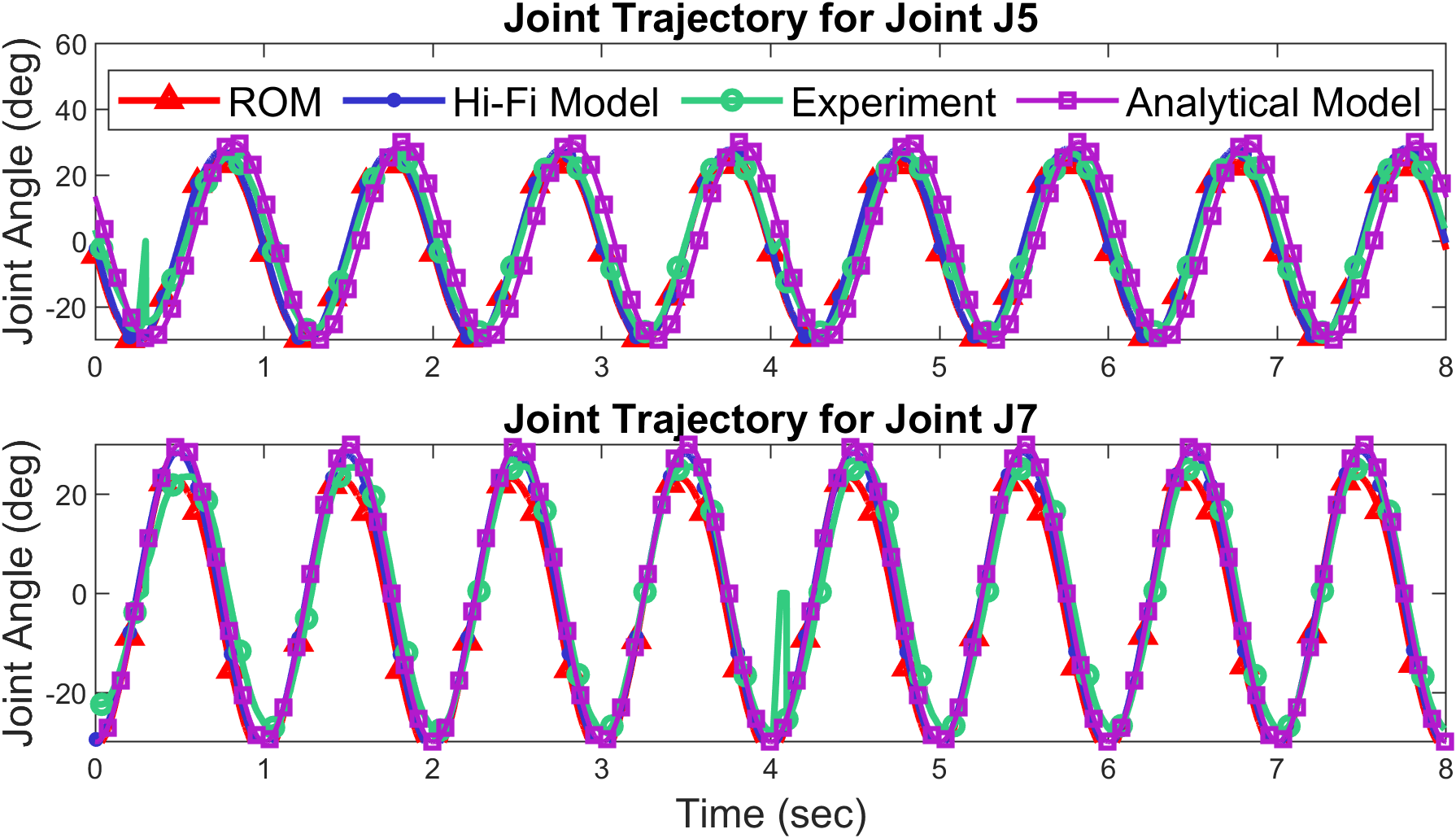}
    \caption{Shows the joint trajectory for two joints during the vertical undulation gait}
    \label{fig:joint_traj}
\end{figure}

\begin{figure}
    \centering
    \includegraphics[width=1\linewidth]{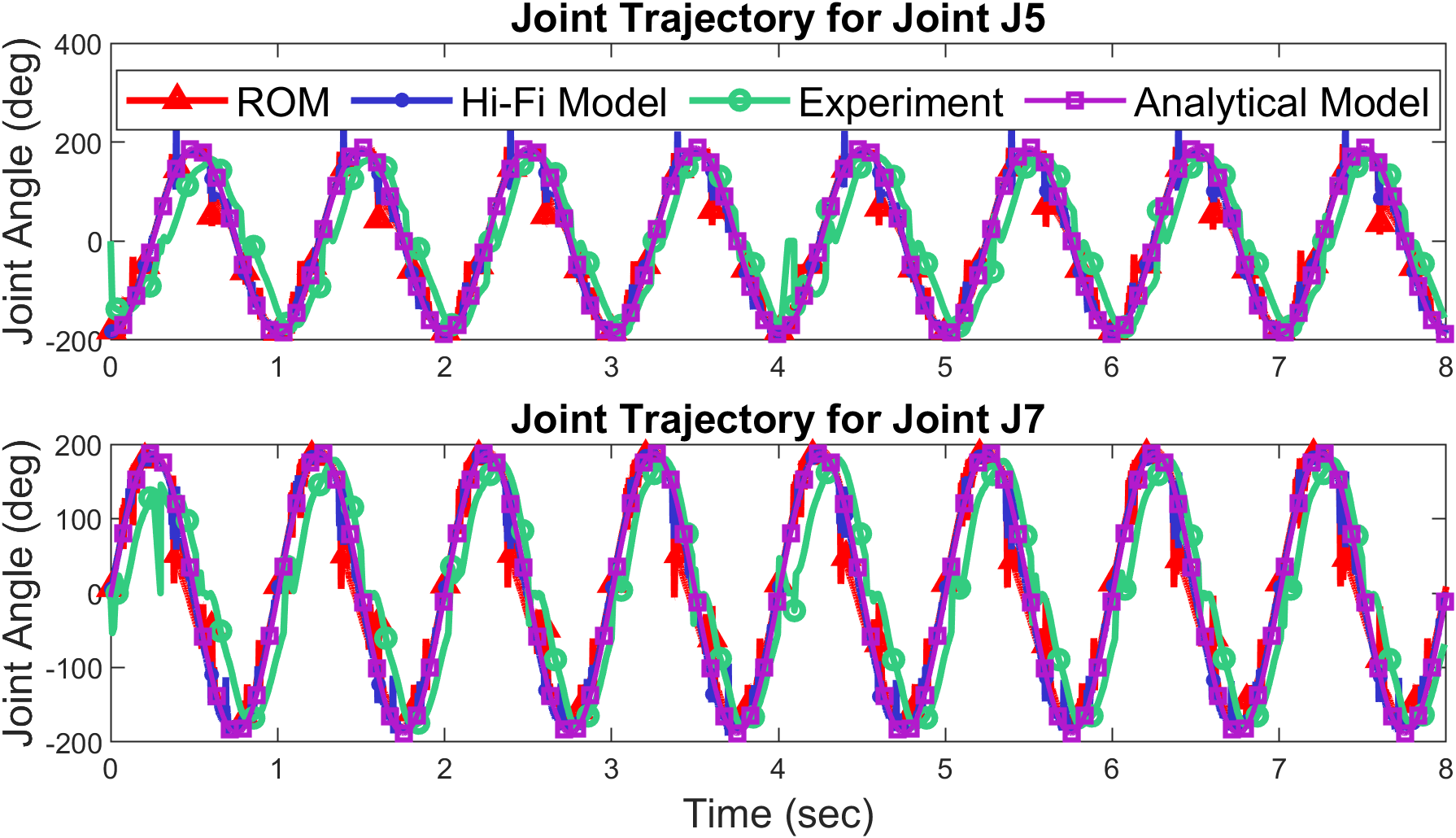}
    \caption{Shows the joint velocity for two joints during the vertical undulation gait}
    \label{fig:joint_vel}
\end{figure}

\begin{figure}
    \centering
    \includegraphics[width=1\linewidth]{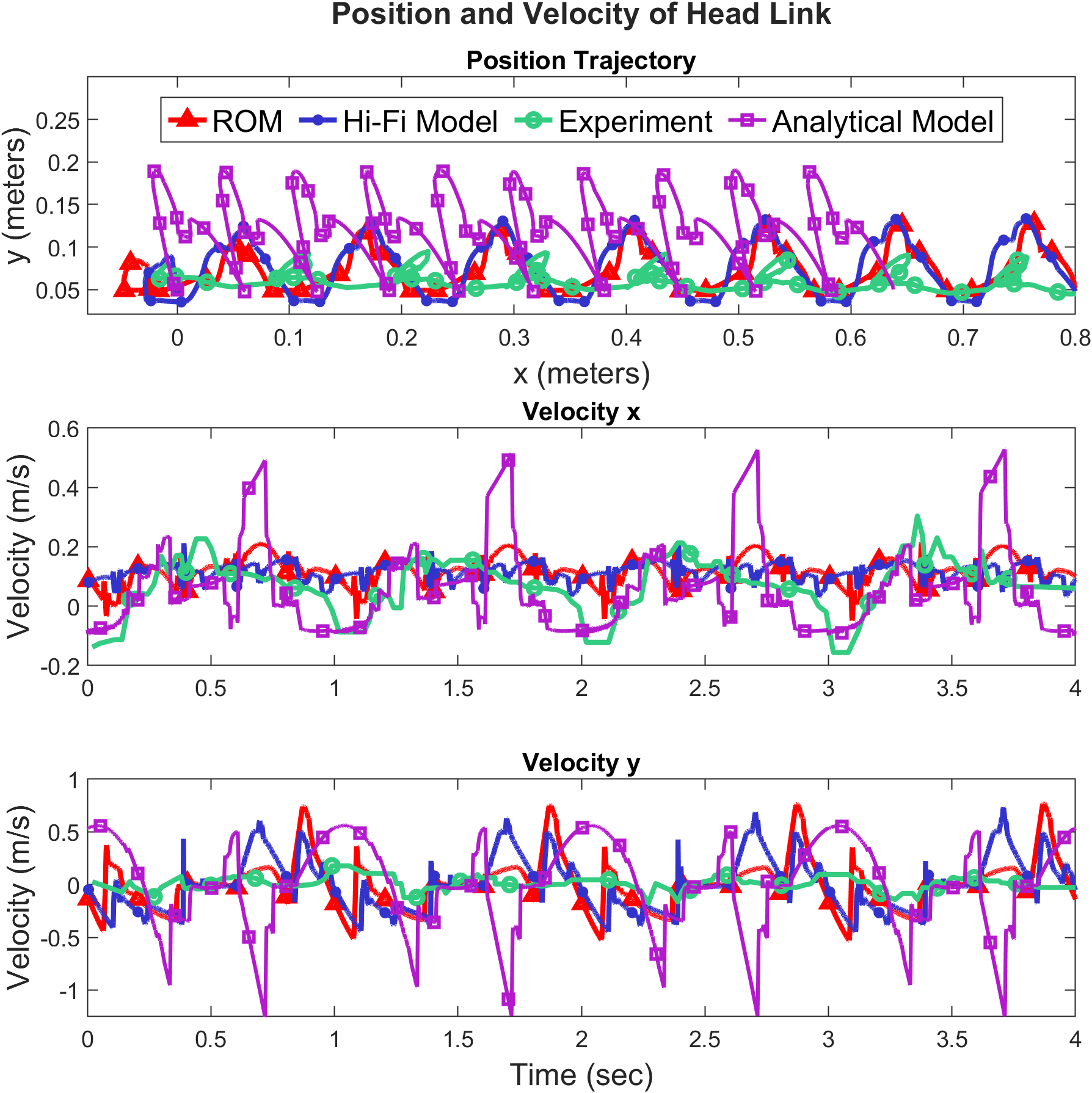}
    \caption{Shows the trajectory of the head module and velocity profile of the robot across the reduced-order model (ROM), high-fidelity simulation, and hardware experiment. The physical robot exhibits a lower vertical velocity and shorter travel distance due to joint backlash, however there is still reasonable agreement between the models.}
    \label{fig:states}
\end{figure}

Figure \ref{fig:snapshots} presents snapshots from the two Simulink simulations alongside experimental results. Over the 10-second trajectory, all models exhibit approximately 1 meter of forward motion. Figures \ref{fig:joint_traj} and \ref{fig:joint_vel} show the position and velocity trajectories of two representative joints, while Figure \ref{fig:states} illustrates the head module’s trajectory and velocity profile across all models and the experiment. The ROM and high-fidelity simulation exhibit strong agreement in behavior. However, discrepancies arise in hardware experiments due to joint backlash, which attenuates joint movements and reduces travel distance. The analytical model predicts larger oscillations with shorter travel distances, likely due to differences in ground reaction force estimation stemming from optimization constraints and tunable hyperparameters, including ground friction and restitution coefficients.

\begin{figure}
    \centering
    \includegraphics[width=1\linewidth]{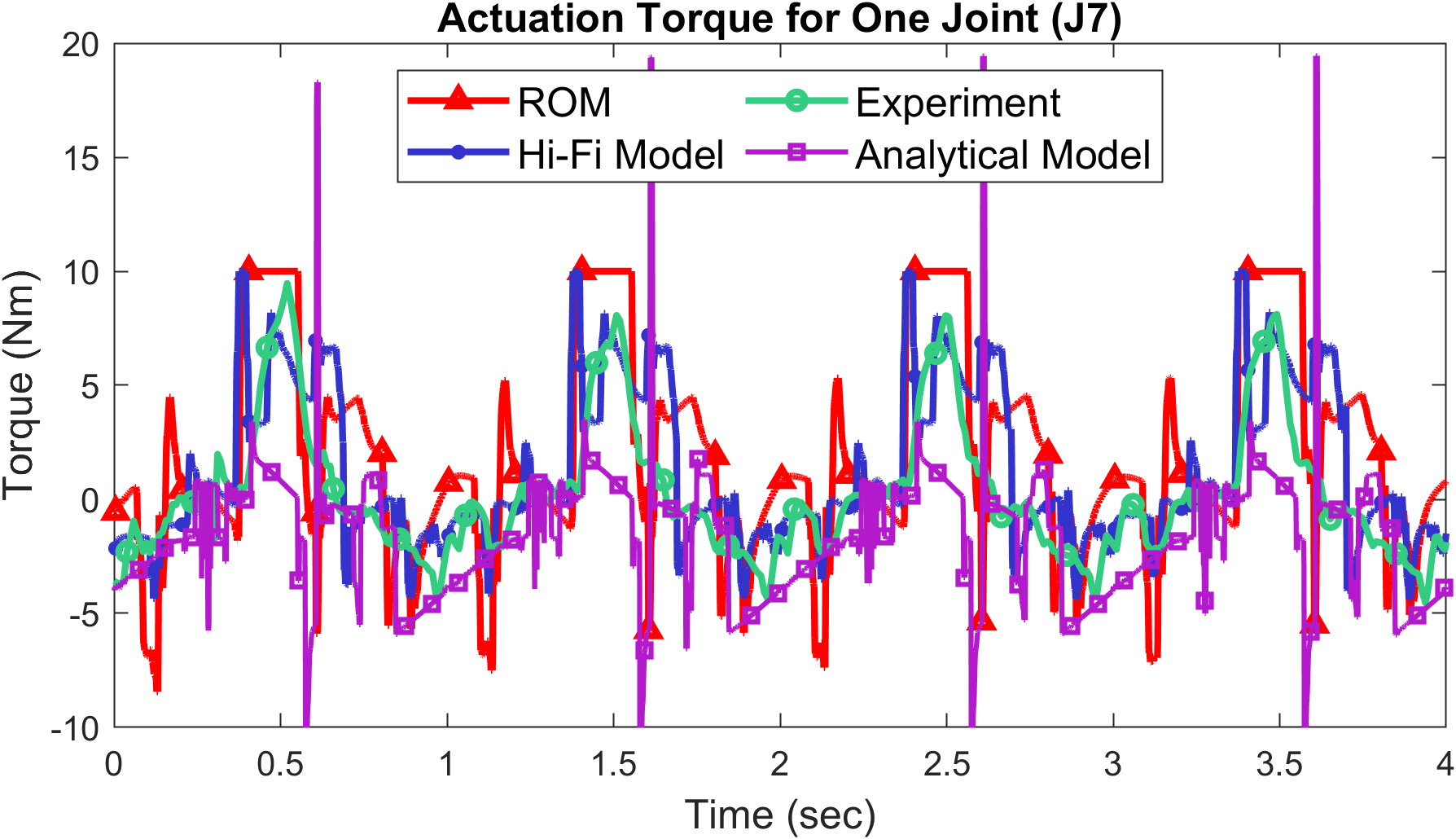}
    \caption{Shows the actuation torque profiles for the ROM, high-fidelity simulation, and hardware experiment. The ROM and fidelity model torques are saturated to 10Nm matching the actuation limits on the hardware platform. The results show strong agreement.}
    \label{fig:inputs}
\end{figure}

The Simulink models approximate ground interactions via a smooth spring-damper system with stick-slip friction opposing the relative velocity of contacting bodies. The spring stiffness and damping coefficient are set to $10^4$ and $10^3$, respectively, with static and dynamic friction coefficients of 0.5. Despite its discrepancies, the analytical model accurately captures the overall trajectory shape and movement direction, with a similar agreement observed in the predicted velocity of the head module.  Figure \ref{fig:inputs} compares actuation torques across models, showing a close match, though the analytical model exhibits sharp torque spikes due to occasional numerical integration instabilities in the optimization.

\begin{figure*}
    \centering
    \includegraphics[width=1\linewidth]{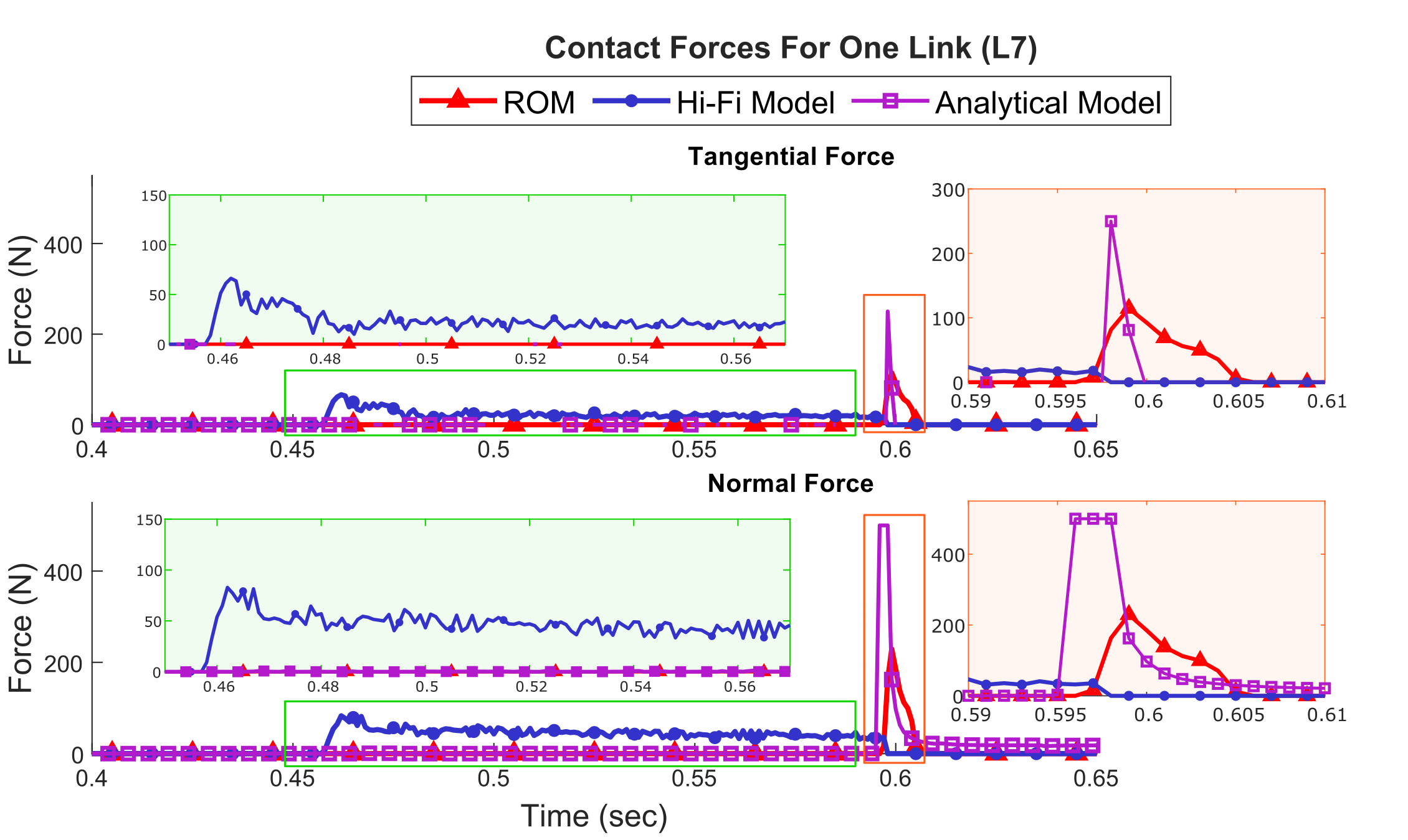}
    \caption{Comparison of ground contact forces between the ROM and high-fidelity model. The ROM exhibits intermittent point contacts, whereas the high-fidelity model demonstrates distributed contact over a longer duration.}
    \label{fig:contact-forces}
\end{figure*}

Figure \ref{fig:contact-forces} compares contact forces for a representative link over one contact period. The ROM and analytical model, which employ point contacts, exhibit shorter contact durations, aligning with the observation that the high-fidelity model, which distributes contact over a larger surface, results in longer contact periods. This distinction manifests as a sharp force spike in the reduced-order models compared to the extended, lower-magnitude contact force in the high-fidelity model. The analytical model predicts higher peak contact forces over shorter durations than the Simulink models.

These findings highlight the effectiveness of the proposed modeling approaches in capturing the robot’s motion dynamics. While the overall trends align well with experimental observations, certain discrepancies—such as joint backlash in hardware and numerical instability in the analytical model—motivate further refinement.

In the following section, we discuss the primary contributions of this work, its limitations, and potential avenues for future improvements.

\section{concluding remarks}
\label{sec:conclusion}
In this work, we develop and validate a reduced-order model (ROM) to analyze the vertical undulation gait of Cobra on flat ground. By representing the robot’s geometry as a series of point masses enclosed within virtual spherical contact regions, the model provides a computationally efficient framework for capturing the core dynamics of undulatory locomotion. Leveraging proximal optimization within a second-order cone program (SOCP), it ensures dynamic feasibility while incorporating realistic ground reaction forces and contact interactions.

While further refinement is needed to improve predictive accuracy, this model establishes a foundation for optimal gait generation in snake robots. Future work will focus on extending the model to gaits with three-dimensional contact patterns, integrating sensor feedback—such as joint torque and inertial measurements—to enhance prediction accuracy, and incorporating terrain mapping using COBRA’s stereo camera. By estimating the gap function from mapped terrain data, the model can be used in a closed-loop framework for autonomous gait and trajectory planning. This would enable optimal contact planning, allowing the robot to autonomously generate joint trajectories for locomotion and loco-manipulation tasks.

\section{ACKNOWLEDGMENT}
This work is supported in part by the U.S. National Science Foundation (NSF) under CAREER Award No. 2340278 and CMMI Grant No. 2142519.

\balance
\printbibliography
\end{document}